\documentclass[11pt,a4paper]{article}
\usepackage{times,latexsym}
\usepackage{url}
\usepackage[T1]{fontenc}

\usepackage{graphicx}
\usepackage{booktabs, multirow}
\usepackage{multirow}
\usepackage{euflag}
\graphicspath{ {./figures/} }
\usepackage{enumitem}

\usepackage[acceptedWithA]{tacl2021v1}

\usepackage{tacl2021v1}

\usepackage{xspace,mfirstuc,tabulary}

\newif\iftaclinstructions
\taclinstructionsfalse 
\iftaclinstructions
\renewcommand{\confidential}{}
\renewcommand{\anonsubtext}{(No author info supplied here, for consistency with TACL-submission anonymization requirements)}
\newcommand{\instr}
\fi

\iftaclpubformat 

\else

\fi

\title{CreoleVal: Multilingual Multitask Benchmarks for Creoles}

\author{
    Heather Lent$^1$,  
    Kushal Tatariya$^2$, 
    Raj Dabre$^3$, 
    Yiyi Chen$^1$, 
    Marcell Fekete$^1$,   \\
    \textbf{Esther Ploeger}$^1$,
    \textbf{Li Zhou}$^{4,7}$,
    \textbf{Ruth-Ann Armstrong}$^{5}$, 
     \textbf{Abee Eijansantos}$^{8}$, \\
     \textbf{Catriona Malau}$^{6}$, 
    \textbf{Hans Erik Heje}$^1$, 
    \textbf{Ernests Lavrinovics}$^1$, 
    \textbf{Diptesh Kanojia}$^9$, \\
    \textbf{Paul Belony}$^{10}$,  
     \textbf{Marcel Bollmann}$^{11}$, 
     \textbf{Loïc Grobol}$^{12}$, 
     \textbf{Miryam de Lhoneux}$^2$, \\
     \textbf{Daniel Hershcovich}$^4$,   
     \textbf{Michel DeGraff}$^{13}$,  
     \textbf{Anders Søgaard}$^4$, 
     \textbf{Johannes Bjerva}$^1$  \\
$^1$Aalborg University, Denmark, $^2$KU Leuven, Belgium, \\
$^3$National Institute of Information and Communications Technology, Japan, \\
$^4$University of Copenhagen, Denmark, $^5$Meta, USA, $^6$University of Newcastle, Australia, \\
$^7$University of Electronic Science and Technology of China, China, \\
$^8$ Zamboanga State College of Marine Sciences and Technology, Philippines, \\
$^9$University of Surrey, UK, 
$^{10}$Kean University, USA, $^{11}$Linköping University, Sweden, \\ $^{12}$Université Paris Nanterre, France, 
$^{13}$Massachusetts Institute of Technology, USA \\
\texttt{hcle@cs.aau.dk}
}

\date{}

\begin{document}
\maketitle

\begin{abstract}
Creoles represent an under-explored and marginalized group of languages, with few available resources for NLP research.
While the genealogical ties between Creoles and a number of highly-resourced languages imply a significant potential for transfer learning, this potential is hampered due to this lack of annotated data.
In this work we present \textsc{CreoleVal}, a collection of benchmark datasets spanning 8 different NLP tasks, covering up to 28 Creole languages; it is an aggregate of novel development datasets for reading comprehension, relation classification, and machine translation for Creoles, in addition to a practical gateway to a handful of preexisting benchmarks. 
For each benchmark, we conduct baseline experiments in a zero-shot setting in order to further ascertain the capabilities and limitations of transfer learning for Creoles. 
Ultimately, we see \textsc{CreoleVal} as an opportunity to empower research on Creoles in NLP and computational linguistics, and in general, a step towards more equitable language technology around the globe. 
\end{abstract}

\section{Introduction}

Despite efforts to extend advances in Natural Language Processing (NLP) to more languages, Creoles are markedly absent from multilingual benchmarks. 
As such, progress towards reliable NLP for Creoles remains impeded, and consequently there is a dearth of language technologies available for the hundreds of millions of people who speak Creoles around the world. 
The omission of Creoles from such benchmarks can be attributed to two key factors: modality and stigmatization. 
The first, modality, is a notable factor as some Creoles are rarely used in writing, and thus text-based NLP is largely moot, highlighting a need for efforts in speech technology for Creoles. 
The latter, stigmatization, is perhaps the most salient of the two, however. 
As the history of many Creole languages is intricately interwoven with broader Western imperialism, colonialism, and slavery, Creole languages are often subjected to the stigmas and prejudices stemming from these historical atrocities \cite{alleyne1971acculturation, degraff2003against}.

On the surface, social prejudices against Creoles may seem extraneous in the context of NLP. 
However, the consequences of this stigmatization are palpable in preventing data collection for these languages. 
For example, it can be greatly challenging to collect data for a language without official status in a given country, even if it is the most widely used language by the populace; common sources for language data like government documentation, educational materials, and local news may not be available.
Moreover, even if a Creole is someone's primary language, sociolinguistic barriers\footnote{In some Creole-speaking communities, the local Creole language is viewed as a ``corrupted'' language, with names like ``broken English''. Thus, speakers of Creoles might not even identify their variety as a separate language.}
rooted in stigma may further prevent people from using it in various contexts, making opportunities for gathering data even more sparse.
Lastly, even when financial resources are available to compensate crowd-workers, logistical challenges can significantly impede data collection efforts for Creole languages \cite{hu-etal-2011-value}.

Stigmatization of Creoles is also an ongoing issue in the scientific domain, which further inhibits work in NLP. 
Indeed, this prejudice is deeply ingrained in linguistics, manifested in the common misconception that Creoles are incomplete or under-developed languages, in direct opposition to concepts like linguistic relativism and Universal Grammar \cite{degraff2005linguists, kouwenberg2009handbook, Aboh2016ANT}.  
This \textit{Othering} of Creoles which has occurred in linguistics has led to a research landscape where Creoles are typically categorized as \textit{exceptions} amongst languages, and thus separated from other languages.
Take, for example, the widely-used WALS database \cite{wals}, which lists Creoles as having the language family ``other''; 
works in NLP or computational linguistics relying on WALS to sample languages from diverse range of families as a part of their methodology consequently exclude Creoles from their work \cite{rama-kolachina-2012-good, vylomova-etal-2020-sigmorphon, bjerva-etal-2020-sigtyp, vastl-etal-2020-predicting, yu-etal-2021-language, chronopoulou-etal-2023-language}.\footnote{For a critical overview of typologically diverse sampling based on language families, see \citet{ploeger2024typological}.}
Beyond WALS, this pattern of exclusion is palpable across NLP, as demonstrated by the marked absence of Creoles in works investigating multilinguality through the lens of language families \cite{majewska-etal-2020-manual, jayanthi-pratapa-2021-study, 10.1162/coli_a_00425, xu-etal-2022-cross}. And while other resources exist to specifically cater to Creoles (e.g., APICS;~\citealp{apics}), the creation of \textit{separate} resources to specifically accommodate Creoles is emblematic of their ghettoization within scientific spaces. In this vein, though Creoles are the singular focus of this work, our datasets, code, and models will allow others to easily incorporate Creoles into broader variety of projects, thus helping remedy the isolation of Creoles across NLP. 

\begin{figure}[t]
\includegraphics[width=1\linewidth]{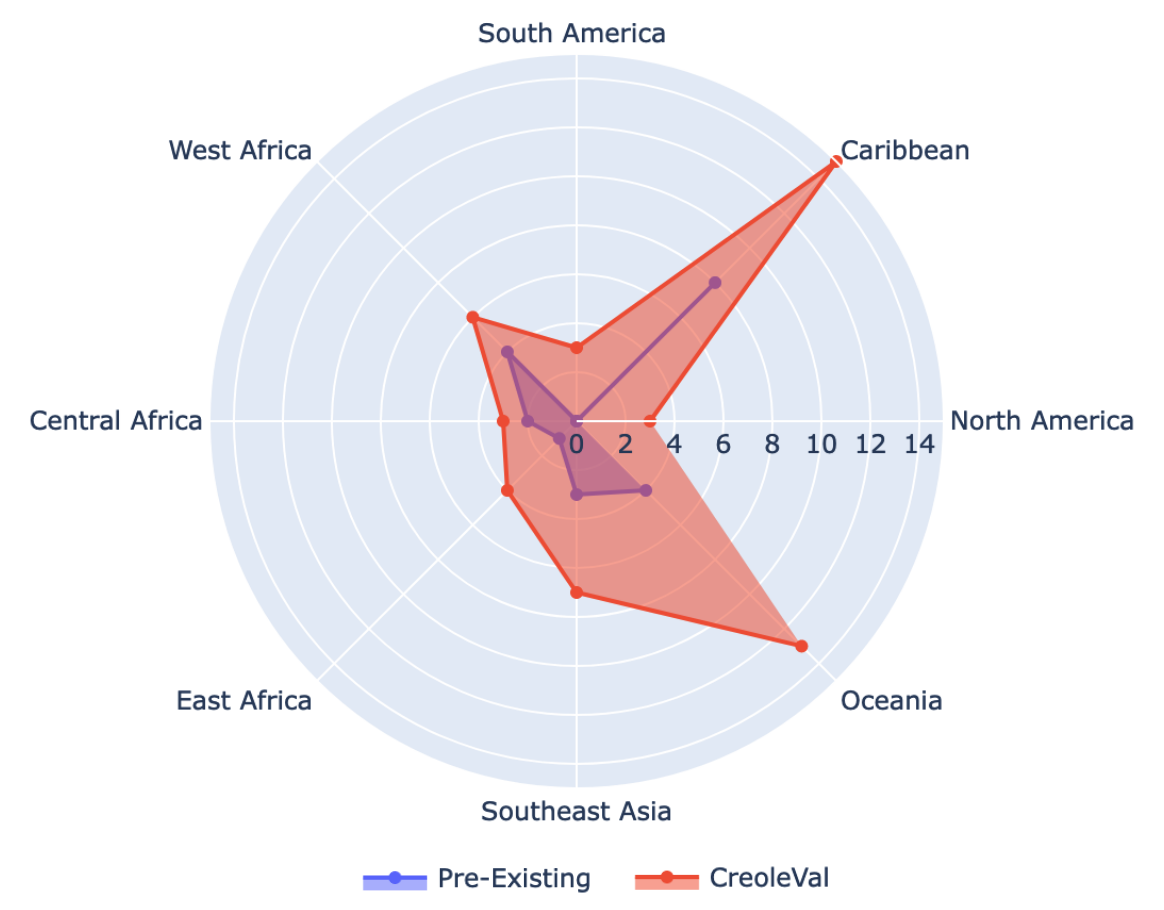}
\caption{\textsc{CreoleVal} expands the availability of labeled data for Creoles around the globe. This chart shows the increased availability of datasets for concrete tasks, across Creoles from different regions. Before \textsc{CreoleVal}, only 11 Creoles had data for at least 1 pre-existing task, and now 28 Creoles have labeled data for at least 1 task and at most 6 tasks. }
\label{fig:map}
\end{figure} 

\paragraph{Inclusion of Creoles}
In an effort to enable NLP research on Creoles, we introduce \textsc{CreoleVal}, a set of benchmarks covering a wide variety of tasks for up to 28 Creole languages.
Enabling NLP research on Creoles offers significant possibilities.
First, this will enable development of language technologies for Creoles, potentially improving technological inclusion of the speakers of these languages.
While increasing the number of NLP datasets for Creoles is important, a crucial note here is that as set of languages, Creoles are not a monolith. 
In some contexts, a Creole can be someone's mother tongue, and the sole language they speak; in other
cases, Creoles can play an important role as a lingua-franca within linguistic diverse communities, and for this reason, deserve special attention of the NLP community \cite{bird-keynote-emnlp-2021}.
Due to their status as marginalized\footnote{Notably, a handful of Creoles do have official language status by law in their respective lands: Haitian Creole, Seychelles Creole, Bislama, and Sango.}
languages, 
we highlight the importance of community involvement when designing \textsc{CreoleVal}.
Inspired by recent recommendations on participatory machine learning \cite{sloane2022participation}, 
we build on previous work by \citet{lent-etal-2022-creole}, and attempt to strike a balance
by creating resources that can be beneficial for both Creole-speaking communities and the NLP community.
Creating the technologies explicitly sought after by various Creole-speaking communities remains an open area for future work, and we believe that the benchmarks and baselines in \textsc{CreoleVal} can be useful to this end.
Second, from a scientific perspective, we argue that Creoles offer an opportunity for careful development and evaluation of transfer learning methods, e.g., leveraging similarities to a Creole's ancestor languages.
For example, consider Chavacano, a language spoken in the Philippines with genealogical ties to Spanish, Tagalog, and other languages.
Below is a sample sentence \citep{apics-46} in Chavacano, with an accompanying Spanish and English translation, annotated with Subject, Verb, and Object roles:
\begin{itemize}[leftmargin=*]
    \item Chavacano: ``Ya-mirá$_V$ el mga ómbre$_S$ un póno de ságing$_O$.''
    \vspace{-3mm}
    \item Spanish: ``Los hombres$_S$ vieron$_V$ un árbol de plátano$_O$.''
    \vspace{-3mm}
    \item English: ``The men$_S$ saw$_V$ a banana tree$_O$.''
\end{itemize}
While Chavacano shares some vocabulary with Spanish, it grammatically maintains the VSO word order of Tagalog. 
Hence, from a transfer learning perspective, one could expect that transfer from Spanish could be useful in terms of lexical overlap, but not syntax.
As many Creoles are genealogically related to higher-resourced languages (e.g. English, French, Spanish, Portuguese, Dutch), resource availability permits research on Creoles that can help shed light on the underlying mechanics of transfer learning.
To this effect, the baselines presented in this work pertain to zero-shot transfer learning, in order to ascertain the current viability of transfer learning for Creoles, in line with previous works for other truly low-resource languages \cite{ebrahimi-etal-2022-americasnli, snaebjarnarson-etal-2023-transfer}. 
Ultimately, the goal of \textsc{CreoleVal} is to facilitate research on transfer learning, computational linguistics, as well as general linguistic research on Creole languages. 
By providing this resource, we hope that inclusion of Creoles in multilingual evaluations will become a default practice in NLP.

\paragraph{Contributions}
In this work, we introduce new datasets for three different NLP tasks (reading comprehension, relation classification, and machine translation) for understudied Creole languages. 
We expand the scope of \textsc{CreoleVal} by packaging these datasets together with pre-existing tasks for Creoles (i.e., dependency parsing, named entity recognition, sentiment analysis, sentence matching,  natural language inference, and machine translation), in a public repository (see Appendix~\ref{appendix:overview} Table~\ref{tab:datasetStats}).  
This repository facilitates further work on Creoles for the NLP community, as we provide a single gateway to this diverse group of languages, allowing for straight-forward data exploration, experimentation, and evaluation. 
The 28 Creole languages covered in \textsc{CreoleVal} are, unfortunately, unequally represented across  tasks due to the difficulties of gathering and curating data. 
However, the addition of our new development data greatly expands upon the existing number of NLP tasks for Creoles (see Figure~\ref{fig:map}). 
For all the datasets comprising \textsc{CreoleVal}, we present baseline experiments with additional analysis on the efficacy of transfer learning for Creoles. 
Our code, data, documentation, and models are available at the public repository.\footnote{\url{https://github.com/hclent/CreoleVal}}
Where we cannot provide data for copyright reasons (i.e., Bible data), we provide detailed documentation and code to allow for reproducibility.

\section{Background}\label{sec:background}

\paragraph{Linguistic Context} The "Creole" label has been assigned to languages known to have arisen through contact between a linguistically diverse set of languages, as a consequence of human movement throughout history \cite{kouwenberg2009handbook}. 
For example, in contrast with Romance languages, which have a clear traceable history from Vulgar Latin \cite{alkire2010romance}, the phylogenetic origins of any given Creole language is more varied. 
This is because Creoles descend from a combination of languages belonging to \textit{different} families \cite{apics}, as illustrated by Creoles across the Caribbean (e.g., Jamaican Patois), which have close ancestral ties to both Indo-European languages (e.g., English) and African ones (e.g., Twi), as a result of European colonialism \cite{patrick2004jamaican}. 
Due to this genealogical context, linguists have looked to Creoles to investigate the process by which new languages emerge \cite{bickerton1983creole, baker1994creativity, mufwene1996founder, lefebvre2001relexification, degraff2001origin, veenstra2008creole} and continue to evolve \cite{croft2000explaining, mufwene2008creoles, mufwene2009evolution, mufwene2015emergence}. 
Among linguists today, there is no consensus on whether Creoles constitute a separate language family \cite{bakker2011creoles, jbp:/content/journals/10.1075/jpcl.31.2.07abo}, or whether the \textit{label} of "Creole" itself is even linguistically valid for discriminating between languages, beyond mere sociohistorical backgrounds \cite{degraff2005linguists, mcwhorter2005defining}.

\paragraph{Previous Work}
Prior works in NLP primarily focus on individual Creole languages, such as 
Antillean Creole \cite{mompelat-etal-2022-parse},
Chavacano \cite{eijansantos-etal-2022-zamboanga},
Jamaican Patois \cite{armstrong-etal-2022-jampatoisnli},
Mauritian Creole \cite{dabre-sukhoo-2022-kreolmorisienmt}, 
Nigerian Pidgin \cite{Ogueji2019PidginUNMTUN, caron-etal-2019-surface, Oyewusi2020SemanticEO, adelani-etal-2021-masakhaner, https://doi.org/10.48550/arxiv.2201.08277, muhammadSemEval2023}, 
Singlish \cite{wang-etal-2017-universal, liu-etal-2022-singlish},
and Sranan Tongo \cite{zwennicker2022general}.\footnote{See \url{https://creole-nlp.github.io/} for a comprehensive list of datasets for Creoles.}
A few works specifically investigate Creoles as a collection of languages, with interest in LMs \cite{lent-etal-2021-language} and transfer learning \cite{lent-etal-2022-ancestor}.
\citet{lent-etal-2022-creole} further discusses some of the social considerations for responsible NLP for Creoles, due to the languages' stigmatization and vulnerability \cite{alleyne1971acculturation, siegel1999stigmatized, kouwenberg2009handbook}. 
We expand upon this existing body of research on Creoles by contributing high-quality evaluation data across a variety of tasks, ensuring that future works in Creole NLP have increased opportunities for measuring progress.
While benchmarking constitutes only a small part of quality assurance for any model in practice, 
the creation of benchmarks also serves as an invitation to the broader research community to engage with new tasks and languages, as evidenced by the success of datasets like MasakhaNER \cite{adelani-etal-2021-masakhaner} and shared tasks \cite{mager-etal-2021-findings, ebrahimi-etal-2023-findings, muhammad-etal-2023-semeval, pal-etal-2023-findings} at bringing more languages into the mainstream of NLP research.
As such, the \textsc{CreoleVal} evaluation benchmarks can similarly 
encourage increased involvement of Creoles in research, with 
the end result of faster progress towards better language technologies for Creole language speakers.

\paragraph{Transfer Learning}
Transfer learning is the process by which a model is trained to make use of knowledge learned in the context of one task or language, with the aim of generalizing to other tasks or languages \textit{outside} the scope of the original training data \cite{Zhuang2019ACS}. 
Over the years, many different techniques have been proposed for achieving cross-lingual transfer, such as learning alignments between words \cite{yarowsky-etal-2001-inducing, DBLP:journals/corr/PadoL14, agic-etal-2016-multilingual, dou-neubig-2021-word} and word vectors \cite{klementiev-etal-2012-inducing, DBLP:journals/corr/abs-1805-11222, kementchedjhieva-etal-2019-lost}, so knowledge from one language can be lent to another on the basis of inferred common ground. Another popular approach relies on unsupervised pre-training of LMs over large corpora, in order to establish a strong but generalized baseline of knowledge \cite{Raffel2019ExploringTL}. In this setting, transfer learning has been effective for extending models trained over higher-resourced languages to lower-resourced ones, especially when the languages in question have similar genealogy, typology, and script \cite{pires-etal-2019-multilingual,wu-dredze-2019-beto,nooralahzadeh-etal-2020-zero,zhao-etal-2021-inducing,de-vries-etal-2021-adapting,de-vries-etal-2022-make}.
In the context of Creoles however, some initial research suggests that transfer-learning from genealogically related languages may not be entirely straightforward.
\citet{de-vries-etal-2022-make} investigate the most effective language pairs for transfer learning of part-of-speech (POS) tagging; while this work does not outright focus on Creoles, a notable finding is that Swedish -- not English nor Portuguese -- is the most useful language for transferring POS tags to Nigerian Pidgin. 
Moreover, in a direct investigation of transfer learning for Creoles, \citet{lent-etal-2022-ancestor} found that LMs trained on multiple ancestor languages failed to transfer well to Creoles on limited downstream tasks. 
Further investigation is required to understand why both the aforementioned studies obtained seemingly counter-intuitive results. 
However, other work investigating the underlying mechanisms that allow for transfer learning have indicated that its success in this setting may be less dependent on genealogical language relatedness, and more dependent on other factors like sub-word overlap \cite{pelloni-etal-2022-subword}.

\paragraph{Multilingual Language Models}
Selecting a pertinent LM is typically the first step for any attempt at transfer learning.
Creoles, however, are largely absent from the most commonly used multilingual LMs (see Table~\ref{tab:mlms}). 
For this work, we choose to work with mBERT \cite{devlin-etal-2019-bert}, XLM-Roberta \cite{conneau-etal-2020-unsupervised}, mT5 \cite{xue-etal-2021-mt5} for natural language understanding tasks, and mBART-50 \cite{DBLP:journals/corr/abs-2008-00401} for generation tasks. 
Despite a lack of coverage for Creoles, these models do include relevant pre-training data for some genealogically related languages.

\begin{table}[tb]
\resizebox{\columnwidth}{!}{%
\setlength{\tabcolsep}{10pt}
\begin{tabular}{@{}lllll@{}}
\toprule
& \textbf{Data} & \textbf{\#Lang} & \textbf{\#Creole} & \textbf{\#Anc} \\
\midrule
\textbf{mBERT} & Wikipedia         & 104              & 1                  & 6                    \\
\textbf{XLM-R} & CC100             & 100              & 0                  & 6                    \\
\textbf{mT5}   & CC4               & 101              & 1                  & 6                    \\
\midrule
\textbf{mBART-50} & custom & 50 & 0 & 5   \\
\bottomrule
\end{tabular}
}
\caption{Coverage of total \textbf{Lang}uages, \textbf{Creole}s, and their \textbf{Anc}estor languages in training data for popular multilingual LMs.
mBERT training data includes Haitian Creole.
For mT5, 0.33\% of the training data comes from Haitian Creole. mBART-50 is trained on the same 25 languages from XLM-R and an additional 25 languages from regular mBART \cite{liu-etal-2020-multilingual-denoising}. While we do not experiment with BLOOM \cite{workshop2023bloom}, it can be noted that 0.0002\% of the Big Science Corpus contains Lingala, a Creole related to Bantu.
}
\label{tab:mlms}
\end{table}

\section{Natural Language Understanding of Creoles}\label{sec:nlu}

Tasks across natural language understanding (NLU) test a model's capacity for grasping syntax and semantics. 
Typical tasks, such as sentiment analysis and named entity recognition, require sizeable amounts of training data for models to exhibit decent performance. 
In order to expand on the availability of NLU data for Creoles we introduce two novel benchmark datasets for reading comprehension and relation classification, before experimenting with a set of pre-existing NLU tasks for Creoles. 
Our baselines are in a zero-shot transfer learning setting for Creoles, as this is the most typical setup for working with languages with little to no data \cite{ebrahimi-etal-2022-americasnli}.

\subsection{Reading Comprehension}
Most pre-existing NLU tasks for Creoles largely examine syntax (see Section~\ref{pre-nlu}), and there is a dearth of NLU tasks for Creoles that evaluate semantic understanding. 
As curating naturally occurring language data for a new task is often prohibitively expensive, dataset translation is a typical alternative, though translation can be complicated by cultural differences between the source and target audience \cite{hershcovich-etal-2022-challenges}. 
In this work, we translate MCTest, a machine reading comprehension dataset introduced by \citet{richardson-etal-2013-mctest}, 
as it pertains to a semantically oriented task, and as the general domain and smaller data size make translation feasible.
Reading comprehension is an NLU task where a model is challenged to correctly answer questions contingent to a specified piece of text. 
The MCTest dataset is composed of short stories intended for school-aged children, each accompanied with four multiple choice questions, that require different levels of reasoning to answer (i.e., context from one or multiple sentences is needed for a human to successfully answer the question).

\paragraph{Translation}
We chose to translate the MCTest160 development set because of the relatively general domain, and smaller size, which makes it feasible for translation (30 stories, 120 questions). 
We hired professional translators, to translate the English MC160 development set, into both Haitian Creole and Mauritian Creole. 
Although we had budget for even more translations, these were the only two Creole languages that we could find professional translators for.  
Notably, there are two different translations for Haitian Creole: a direct translation, and a localized translation. 
As opposed to the direct translation, the localized version is a culturally-sensitive translation, with minor changes to include names, places, and activities that are directly pertinent to a Haitian audience \cite{roemmele2011choice}. 
For example, the original English dataset may discuss an ice cream truck (directly translated to "\textit{kamyon krèm}"), though ice cream is not a typical desert in Haiti; thus in the localized dataset, "ice cream truck" has been changed to "\textit{machann fresko}", a cart which sells a shaved-ice desert enjoyed in Haiti.
The addition of these two different Haitian Creole datasets for reading comprehension, additionally paves the way for evaluating progress in cross-cultural NLP \cite{hershcovich-etal-2022-challenges}. 

\begin{table}[tb]
\centering
\resizebox{\columnwidth}{!}{%
\setlength{\tabcolsep}{19pt}
\begin{tabular}{@{}lrr@{}}
\toprule
& \textbf{mBERT} & \textbf{XLM-R} \\
\midrule
\textbf{Haitian-direct}  &    51.60\%     &         39.16\%     \\
\textbf{Haitian-localized} &    50.83\%    &             43.33\%  \\ 
\textbf{Mauritian}         &    49.10\%     &           43.33\%   \\ 
\midrule
\textbf{English}           &   63.33\%    &           45.00\%     \\ 
\bottomrule
\end{tabular}}
\caption{Accuracy results for MCTest160 development data, when trained on the English MC160 training data.}
\label{tab:mctest}
\end{table}

\begin{table*}[th]
\centering
\resizebox{\textwidth}{!}{%
\setlength{\tabcolsep}{12pt}
\begin{tabular}{@{}llrrrr|rrrr@{}}
\toprule
\multirow{2}{*}{\textbf{Dataset}} & \small\textbf{Sent. Enc.} & \multicolumn{4}{c|}{\textbf{\texttt{bert-base-multilingual-cased}}}                                 & \multicolumn{4}{c}{\textbf{\texttt{xlm-roberta-base}}}                       \\ \cmidrule(l){3-10} 
&\small\textbf{Rel. Enc.} & \textbf{Bb-nli} & \textbf{Bl-nli} & \textbf{Xr-100} & \textbf{Xr-b} & \textbf{Bb-nli} & \textbf{Bl-nli} & \textbf{Xr-100} & \textbf{Xr-b} \\ \midrule

        \textbf{Dev(en)}&  & 59.63±3.48 & 76.15±1.59 & 63.47±1.75 & 62.15 ±1.65 & 46.76±2.58 & 50.58±2.08 & 49.11±2.51 & 49.04±1.49\\
        \midrule  \midrule
    \textbf{bi} & & 28.01±2.42 & 25.61±3.92 & 27.66±5.45 & 25.96±3.80 & 18.81±4.04 & 9.62±0.78 & 19.42±4.51 & 14.79±1.77 \\
        \textbf{cbk-zam} & & 20.06±5.88 & 20.85±6.03 & 17.67±6.68 & 17.39±6.45 & 27.08±6.86 & 18.48±6.83 & 18.50±2.77 & 20.32±2.73 \\ 
        \textbf{jam}&  & 26.97±5.87 &15.65±5.00 & 20.07±5.93 & 23.98±7.24 & 10.62±1.27 & 9.42±5.71 & 9.06 ±1.70 & 10.22±0.92 \\
        \textbf{tpi}&  & 23.57±4.17 & 22.90±2.97 & 22.86±8.13 & 21.42±5.96 & 9.36±3.77 & 11.64±5.54 & 8.31±8.07 & 8.48±4.78 \\ 

        \textbf{AVG} & &	\textbf{24.65}	&	21.25	&	22.06	&	22.19		&\textbf{16.47}	&	12.29	&	13.82	&	13.45\\

\bottomrule
\end{tabular}}
\caption{Relation Classification performance measured by macro F1 score on English validation (dev) set and Creole test sets. AVG shows the overall performance per setup across all Creole languages. \textbf{Bold} indicates the best performance for each sentence encoder setting. Sent. Enc.: sentence encoder. Rel. Enc.: relation encoder.}
\label{tab:overall-results-RE}
\end{table*}

\paragraph{Results and Analysis}
For our benchmark experiments on the Creole MCTest160 development set, we use a simple transformer-based baseline approach, leveraging mBERT and XLMR as the basis of these models. 
We finetune them for 10 epochs over the English MCTest160 training set. 
A summary of our results is in Table~\ref{tab:mctest}, with full results and hyperparameter settings documented in the accompanying Github repository. 
mBERT out-performs XLMR, although XLMR performs better over the localized data than the direct translation for Haitian. 
The performance on Haitian can likely be attributed to the fact that mBERT has been pre-trained on Haitian, while XLMR has not. Meanwhile, the performance on Mauritian is surprising as neither models have seen this language.
It's particularly noteworthy that mBERT results on Creoles outperforms XLMR's English performance by far.
In comparison, a random baseline on MCTest160 yields an accuracy of 25\%, and Attentive Reader \cite{hermann2015teaching} has an accuracy of 42\% on English data.


\subsection{Relation Classification}\label{sec:rel_class}
Relation classification (RC) aims to identify semantic associations between entities within a text, essential for applications like knowledge base completion~\citep{lin2015learning} and question answering~\citep{xu2016question}. In this work, we introduce the first manually-verified RC datasets for four Creole languages: Bislama, Chavacano, Jamaican Patois, and Tok Pisin.

Our dataset is sourced from Wikipedia, where we found 16 Creoles with a presence, though only 9 had readily-available Wikidumps.\footnote{bi, cbk-zam, gcr, hat, jam, pap, pih, sg, tpi} While Wikipedia is a common source for gathering data, poor quality of articles is an outstanding issue known to plague Wikipedias of lower-resourced languages \cite{kreutzer-etal-2022-quality}. 
As such, the creation of our RC datasets involves speakers of the Creoles to ensure quality, and preserve the domain, allowing for integration of Creoles into the broader spectrum of RC projects \cite{sorokin-gurevych-2017-context, koksal-ozgur-2020-relx, nag-etal-2021-data, chen-li-2021-zs, chen-etal-2022-multilingual}. 

To construct the dataset, we first preprocess\footnote{\url{https://github.com/attardi/wikiextractor}} Wikipedia dumps and perform automatic entity linking using OpenTapioca~\citep{delpeuch2019opentapioca}. 
Unsurprisingly, we observe many Creole Wikipedia entries are short and \textit{templatic}, possibly due to machine generation. This templatic nature, however, facilitates the annotation process for RC, as it allows for straight-forward identification of entities and relations by the authors of this work who have linguistic training. 
For example, consider the following examples from the Tok Pisin Wikipedia:
\begin{itemize}[noitemsep,topsep=1pt]
    \item \textcolor{darkblue}{Talin} i kapitol bilong \textcolor{darkblue}{Estonia}.
    \item \textcolor{darkblue}{Vilnius} i kapitol bilong \textcolor{darkblue}{Lituwenia}.
    \item \textcolor{darkblue}{Busares} i kapitol bilong \textcolor{darkblue}{Romenia}.
    \item \textcolor{darkblue}{Budapest} i kapitol bilong \textcolor{darkblue}{Hangri}.
    \item \textcolor{darkblue}{Stockholm} i kapitol bilong \textcolor{darkblue}{Suwidan}.
\end{itemize}
From these samples above, we can infer a latent template of "\textcolor{gray}{[[CITY]] is the capital of [[COUNTRY]]}", with the entities having the relation "capital of" (P1376 in Wikidata). 
Thus, to facilitate manual annotation of relations, and corrections of the automatic entity tagging, we automatically cluster sentences based on the latent templates. Thereafter, sentences with annotated triples not found in Wikidata are discarded.  

After the annotation process, speakers of the pertinent Creole languages assessed the quality of the samples, and furthermore provided spelling and grammar corrections, where deemed necessary. 
This quality assurance process was complemented by a linguistic expert who cross-referenced the datasets with linguistic grammars to identify possible errors.
The process resulted in high-quality evaluation data for 4 of the 9 initially identified Creole Wikipedias, with each dataset contains 97 evaluation samples\footnote{For a complete discussion on dataset creation, latent templates, and manual review processes, see Appendix~\ref{appendix:rel_class}}.

We establish a benchmark for Creole RC using a zero-shot cross-lingual transfer approach: we employ LMs that have not been exposed to the Creoles and train exclusively on English data.

\begin{table*}[th]\centering
\resizebox{\textwidth}{!}{%
\setlength{\tabcolsep}{19pt}
\begin{tabular}{@{}lccrrrrr@{}}
\toprule
\textbf{Task} &\textbf{Language} &\textbf{Dataset} &\textbf{Metric} &\textbf{mBERT} &\textbf{XLM-R} &\textbf{mT5} \\
\midrule
\multirow{2}{*}{\textbf{UDPoS} (supervised)} &pcm &UD\_Naija-NSC \cite{caron-etal-2019-surface} & Acc &0.98 &\textbf{0.98} &0.98 \\ 
&singlish &Singlish Treebank \cite{wang-etal-2017-universal} & Acc &0.91 &\textbf{0.93} &0.91 \\
\midrule
\multirow{8}{*}{\textbf{NER} (supervised)}  &pcm &MasakhaNER \cite{adelani-etal-2021-masakhaner} & Span-F1 &0.89 &0.89 &\textbf{0.90} \\
\cline{2-7} 
&bis &\multirow{7}{*}{WikiAnn \cite{pan-etal-2017-cross} } &\multirow{7}{*}{Span-F1} &\textbf{0.94} &0.90 &0.72 \\
&cbk-zam & &  &\textbf{0.96} &0.96 &0.94 \\
&hat & &  &0.78 &\textbf{0.84} &0.48 \\
&pih & &  &\textbf{0.90} &0.88 &0.61 \\
&sag & &  &0.89 &\textbf{0.93} &0.79 \\
&tpi & &  &\textbf{0.91} &0.89 &0.75 \\
&pap & &  &\textbf{0.90} &0.89 &0.85 \\
\midrule
\multirow{2}{*}{\textbf{SA} (supervised)} &pcm & AfriSenti \cite{muhammadSemEval2023} & Acc & 0.66 &\textbf{0.68} &0.67 \\
&pcm & Naija VADER \cite{Oyewusi2020SemanticEO} & Acc & 0.71 &\textbf{0.72} &0.72 \\
\midrule
\textbf{NLI} (few-shot) &jam &JamPatoisNLI \cite{armstrong-etal-2022-jampatoisnli} & Acc &0.74 &\textbf{0.76} &0.66 \\
\midrule
\multirow{7}{*}{\textbf{Sentence Matching}} & cbk-eng &\multirow{7}{*}{Tatoeba \cite{artetxe-schwenk-2019-massively}} & \multirow{7}{*}{Acc} & \textbf{15.9} & 3.9 & 6.5 \\
& gcf-eng & & & \textbf{12.8} & 4.9 & 6.9 \\
& hat-eng & &  & 23.9 & 18.5 & \textbf{37.9}\\
& jam-eng & &  & \textbf{19.9} & 9.6 & 10.3 \\
(zero-shot) & pap-eng & &  & \textbf{22.4} & 6.1 & 15.9 \\
& sag-eng & &  & 5.7 & 2.1 & \textbf{7.3} \\
& tpi-eng & &   & 7.2 & 3.3 & \textbf{7.6} \\
\bottomrule
\end{tabular}
}
\caption{Baseline scores for pre-existing NLU tasks for Creoles: dependency parsing (UDPoS), named entity recognition (NER), sentiment analysis (SA), natural language inference (NLI) , and sentence matching. Additional experiments, results, and analysis are included in the CreoleVal repository's documentation. }\label{tab:NLU-Baselines}
\end{table*}

\paragraph{Model and Training}
We adopt the method introduced by \citet{chen-li-2021-zs}, which excels in zero-shot transfer learning for RC on Wikipedia and Wikidata~\citep{han-etal-2018-fewrel}. This approach projects both sentences and their associated relation descriptions into a shared embedding space, minimizing distances between them while performing classification. 
For training, we use the UKP dataset~\citep{sorokin-gurevych-2017-context}, which contains 108 Properties (i.e., relations in Wikidata). In contrast, our Creole datasets feature just 13 Properties, four of which are not present in the UKP dataset. Five relations are separated for validation. We fine-tune multilingual mBERT and XLM-R~\citep{conneau-etal-2020-unsupervised} models using multilingual sentence transformers~\citep{reimers-2019-sentence-bert}. The sentence encoder employs mBERT and XLM-R,\footnote{Respectively, \texttt{bert-base-multilingual-cased}, \texttt{xlm-roberta-base}.} while the relation encoder uses one of four alternative models, denoted Bb-nli, Bl-nli, Xr-b, Xr-100\footnote{Respectively,\texttt{bert-base-nli-mean-tokens}, \texttt{bert-large-nli-mean-tokens}, \texttt{xlm-r-bert-base-nli-mean-tokens}, \texttt{xlm-r-100langs-bert-base-nli-mean-tokens}.} here, as sentence embeddings of the relation descriptions from Wikidata.

\paragraph{Results and Analysis}
Table~\ref{tab:overall-results-RE} shows the performance of RC in each setting. 
We observe worse performance in the Creoles than English. 
This highlights the particular challenge of leveraging pretrained LMs for zero-shot cross-lingual transfer for RC for Creoles, due to the lack of representation of Creoles in the LM training data. 
In addition, the choice of the sentence encoder is a primary determinant of performance of Creole RC. When using mBERT as the sentence encoder, the performance of Creole RC tends to be slightly better than XLMR. Under the same sentence encoder, different relation encoders exhibit slight variations in performance. 
We speculate that mBERT may perform better due to its pre-training over Wikipedia, in contrast to XLMR, which is pre-trained over a wider variety of domains. Previous works on multilingual factuality also observe mBERT outperforming XLMR \cite{jiang-etal-2020-x, fierro-sogaard-2022-factual}.

\subsection{Prior NLU Benchmarks}\label{pre-nlu}
In addition to the datasets that we introduce, there are a handful of pre-existing, labeled datasets for Creole languages in the space of NLU. 
In order to facilitate concentrated efforts on Creole NLP, we have gathered these tasks and packaged the baseline experiments for them with the CreoleVal repository. 
For each of these prior benchmarks, we provide code to run baseline experiments with three multilingual LMs (mBERT, XLM-R and mT5). 
In contrast with the brand new datasets presented in \textsc{CreoleVal}, the majority of prior benchmarks allow for supervised learning. 
Thus, in order to ascertain the expected performance for these tasks given the data available, we train and evaluate fully supervised models, where training data exists (UDPoS, NER, and sentiment analysis). 
For JamPatoisNLI \cite{armstrong-etal-2022-jampatoisnli}, we reproduce the authors' results by following the reported methodology: first we fine-tune on English MNLI \cite{N18-1101}, before doing few-shot learning on 250 samples of Jamaican Patois. 
The sentence matching Tatoeba task \cite{artetxe-schwenk-2019-massively} is the only without dedicated training or few-shot data, and thus is the only task where we evaluate the zero-shot performance of the pertinent LMs. 
The performance on the test set for each task and LM in Table \ref{tab:NLU-Baselines}.
Unsurprisingly, performance is best when training data is available, though few-shot learning shows promising results in the case of JamPatoisNLI. 
However, previous work has noted that a high token overlap is needed to successfully achieve cross-lingual transfer for languages \textit{unseen} by a pre-trained LM \cite{winata-etal-2022-cross}. As spelling conventions for many Creoles have greatly diverged from those of ancestor languages (e.g., "Pwofesè" in Haitian Creole to "Professeure" in French), subword token overlap between Creoles and related languages will likely be low, and therefore few-shot learning may not help in such scenarios. 
As additional samples for few-shot learning are  not available for most Creoles, there is an outstanding need for improved zero-shot performance via transfer learning, until further data can be curated.

\section{Natural Language Generation of Creoles}\label{sec:nlg}
Unlike NLU, where the model aims to predict an accurate label, natural language generation (NLG) is arguably a more challenging task as models should generate output that is \textit{adequate} as well as \textit{fluent}.   
A lack of data -- both in terms of size and domain -- further complicates NLG for Creole languages. 
In this paper, we introduce 2 new machine translation (MT) datasets for Creoles.
The first covers 26 Creoles with text drawn from the religious domain, and the second is a small, but very high quality, Hatian Creole dataset in the educational domain. 
We also conduct experiments and evaluate performance on a pre-existing MT dataset for Mauritian Creole.

\subsection{CreoleM2M MT}
As the world's most translated text, the Bible is a typical starting point for gathering language data in a low-resource scenario. 
While Bible data has a number of limitations 
(e.g., fixed domain, archaic language, and translationese \cite{mielke-etal-2019-kind}),
notable benefits include its size and parallelism with other languages, which lends itself aptly to MT.
We gathered parallel corpora for 26 Creole Bibles from \citet{mayer-cysouw-2014-creating},\footnote{To access the raw Bible corpora, one must request the authors due to copyright issues.} along with additional texts from the JW300 corpus \cite{agic-vulic-2019-jw300}.
In total, our parallel MT corpus contains 3.4M sentences and 71.3M and 56.3M Creole and English words, respectively, making it the largest Creole parallel corpus to date.
Furthermore, we split from the Bible part of the corpus, 1,000 and 2,000 sentences for each Creole and English and use them for development and testing, respectively. 
Note that the development and test sets are N-way parallel (N=27: 26 Creoles and English). 
We ensured that there is no overlap between the training, development, and test data. See Appendix~\ref{appendix:mt} for exact details on dataset sizes. 

\subsubsection{Experiments}
We fine-tune mBART-50-MT \cite{DBLP:journals/corr/abs-2008-00401} and also train models mBART from scratch, over the parallel Bible text. 

\paragraph{Vocabulary}
For models trained from scratch, we use the training data and create a shared tokenizer of 64,000 subwords  for all 26 Creoles and English using \textit{sentencepiece} \cite{kudo-richardson-2018-sentencepiece}. 
Due to the large number of languages, we only train bilingual models and leave multilingual models for future work. While we could have created separate vocabularies for bilingual models, a shared tokenizer will be helpful in ensuring consistency with future planned multilingual model experiments. For the fine-tuned models, we use the mBART-50 tokenizer containing 250,000 subwords. Although this tokenizer's vocabulary was not explicitly trained on Creoles, we expect the subwords from related parent languages to be sufficient.

\begin{figure*}[th!]
\centering
\includegraphics[width=0.49\textwidth]{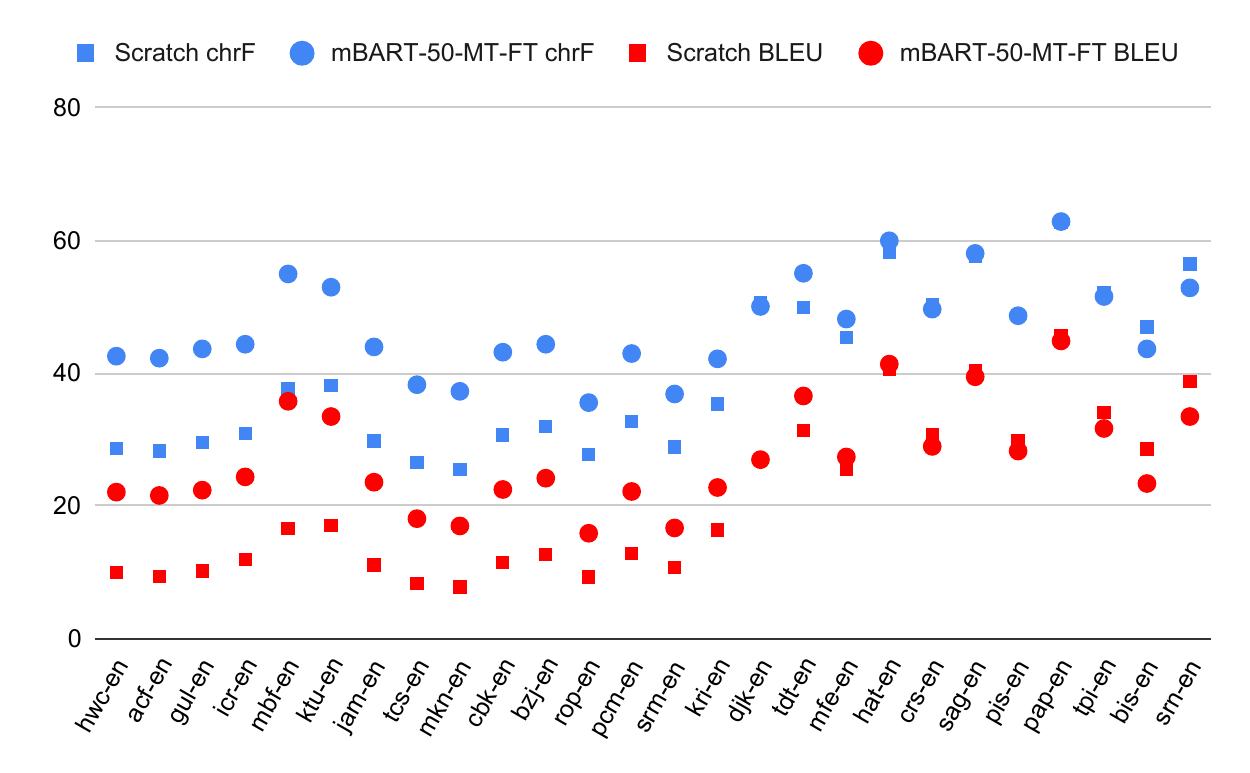}
\includegraphics[width=0.49\textwidth]{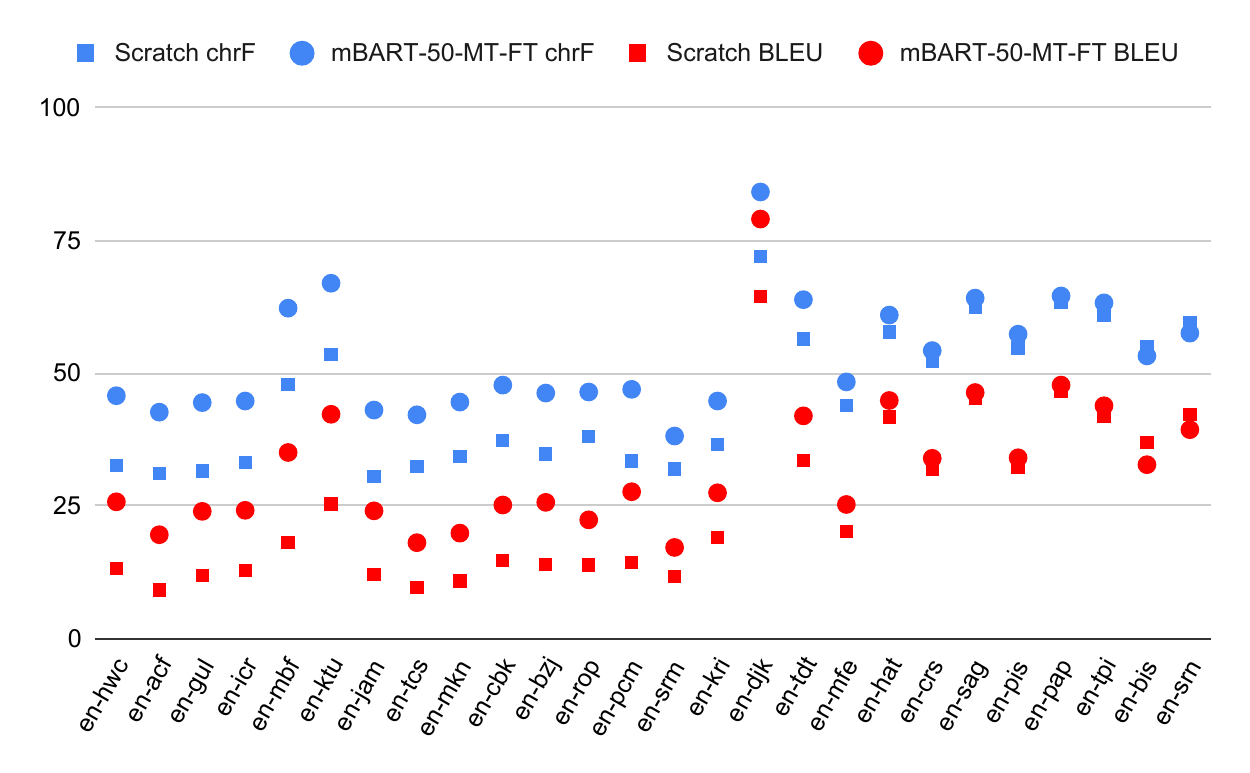}
\caption{chrF (blue color) BLEU (red color) scores obtained using baseline models (scratch; square points) and fine-tuned models (mBART-50-MT-FT; circle points) on the Bible corpus for XX-En (left) and En-XX (right) language pairs, where XX represents Creole languages. The language pairs are ordered from left to right in increasing sizes of parallel corpora from 4,366 pairs to 583,746 pairs}
\label{fig:bleu_chrf_bible}
\end{figure*}

\paragraph{Training}
We trained our models using the YANMTT toolkit\footnote{\url{https://github.com/prajdabre/yanmtt}} \cite{DBLP:journals/corr/abs-2108-11126}, which supports training models from scratch as well as by fine-tuning mBART models. Here, we train models from scratch as well as by fine-tuning the mBART-50-MT model\footnote{\url{https://huggingface.co/facebook/mbart-large-50-many-to-many-mmt}} following~\citet{dabre-sukhoo-2022-kreolmorisienmt}. 
The training utilizes Adam optimizer \cite{kingma2014adam}, and trains till convergence. We evaluate the training performance on the development set using BLEU score after every 1,000 training steps. The training process determines convergence when BLEU scores do not improve for 20 consecutive evaluations.\footnote{Note that we anneal the learning rate by half when the BLEU scores don't improve for 10 consecutive evaluations and then again by half if the scores don't improve for 15 consecutive evaluations. Therefore, after cutting the learning rate by half (each time) for the final convergence decision, we wait for 20 consecutive evaluations to declare model convergence.}

\paragraph{Decoding}
We perform decoding using beam search with a beam of size $4$ and a length penalty of $0.8$. Due to the large number of language pairs, we do not tune these parameters for each language pair.

\paragraph{Results and Analysis} 

Figure~\ref{fig:bleu_chrf_bible} shows the performance in terms of chrF and BLEU scores for Creole to English and English to Creole translation for the test set of the CreoleM2M benchmark. For models trained from scratch, performance appears correlated with the size of the parallel corpus. Therefore, fine-tuning the mBART-50-MT model leads to significant improvements in translation quality by up to 19.2 BLEU and 17.3 chrF for Creole to English translation and up to 16.9 BLEU and 13.5 chrF for English to Creole translation. We noted that both BLEU and chrF scores are correlated\footnote{We calculated a Pearson correlation score of 0.98.} with each other. We note that fine-tuning is not always a good idea for the Creoles with more training data available. In most larger-resourced settings, we observed a drop in translation quality, indicating that the fine-tuned model converges too quickly, and is unable to learn well from the training data.

\subsection{MIT-Haiti MT}
While Bible translations can provide initial data for training MT systems, this domain is markedly limited, highlighting a need for MT datasets for Creoles originating from other, more generalizable domains. 
To this end, we introduce the \textsc{Bank done MIT-Ayiti}, or in English, the MIT-Haiti Corpus: a manually-verified, 
high-quality collection of parallel Haitian Creole sentences with English, French, and Spanish translations.  
This data comes from Platform MIT-Haiti\footnote{\url{https://mit-ayiti.net/}}, a learning platform with educational material for students in Haitian Creole. We scrape the entire website, including the web text and PDFs.  
The parallel sentences for this MT corpus come from 60 multilingual stories (the PDFs and their converted plain text transcriptions);
these stories were each manually cleaned and corrected (i.e., in cases where the PDF-reader made mistakes in transcribing, these were manually corrected), aligned, and verified by a subset of the authors, who have qualifications in both linguistics and NLP. 
For the remaining monolingual Haitian text without direct parallel translations, we manually clean and verify these sentences with the same process, and release a small  set of monolingual examples ($\sim$8200 utterances), which could potentially be useful for few-shot continued pre-training of a language model. 
Although this dataset is relatively small, we would like to stress that it is high quality, as it comes directly from a community that actively fosters education and writing in Haitian Creole.

\paragraph{OPUS for MIT-Haiti}
To establish the baseline performance on the MIT-Haiti Corpus, we leverage pre-trained OPUS-MT models \cite{tiedemann-thottingal-2020-opus}.
In Table \ref{tab:mit-ayiti}, we show the performance of pre-trained OPUS-MT models on the MIT-Haiti benchmarks.
These models were previously benchmarked on the Tatoeba and/or JW300 corpus, which are limited in complexity and domain, respectively. 
By extending this to the MIT-Haiti Corpus, we can gain an insight into the performance of these models on more diverse usage of Haitian Creole.
We translate from Spanish, French and English into Haitian Creole, because this translation direction has the potential to be useful for (monolingual) speakers of Haitian Creole, as it provides increased information access. 
Notably, the scores on the MIT-Haiti benchmarks are considerably lower than those on previous benchmarks. 
For instance, the English to Haitian Creole model scores 45.2 BLEU and 59.2 chrF on the Tatoeba test set\footnote{\url{https://huggingface.co/Helsinki-NLP/opus-mt-en-ht}}, while it retrieves only 14.7 BLEU and 35.8 chrF on the MIT-Haiti Corpus. 
This suggests that previous benchmarks are likely to be overly optimistic.

\paragraph{CreoleM2M for MIT-Haiti}
Table~\ref{tab:mit-ayiti} contains the results for the fine-tuned CreoleM2M models on the MIT-Haiti Corpus. We can see that the BLEU and chrF scores are 18.6/38.1 and 22.0/43.9 for Haitian Creole to English and English to Haitian Creole, respectively. Despite the domain differences between CreoleM2M's training data (religion) and the MIT-Haiti benchmarks (education), a brief manual inspection revealed that the translation quality is not particularly bad, however the generated translations tend to contain spurious religious content. Extensive human evaluation of these translations will help in better understanding of the limitations of our CreoleM2M models in a cross-domain setting.

\subsection{Prior NLG Benchmarks}\label{sec:prior}

\paragraph{KreolMorisienMT} \cite{dabre-sukhoo-2022-kreolmorisienmt} is a dataset for machine translation of Mauritian Creole (i.e., Kreol Morisien) to and from English and French. The dataset spans multiple domains spanning the Bible, children's stories, commonly used expressions and some books. We refer the reader to \citet{dabre-sukhoo-2022-kreolmorisienmt} for further details. In this paper, we focus only on translation to/from English. We combine the training data from the Kreol Morisien part of the CreoleM2M dataset with KreolMorisienMT's training data and then train MT models to show the impact of our newly mined data. We filter out those sentences from CreoleM2M, which are present in the development and test sets of KreolMorisienMT, for clean evaluation. This gives us 188,820 sentence pairs, which is almost an order of magnitude larger than the 21,810 sentence pairs in KreolMorisienMT. As a baseline, we only train models with the CreoleM2M data containing 167,010 sentence pairs after removing the development and test set sentences of KreolMorisienMT.

For the KreolMorisienMT test set, since it is standalone, we focus on standalone bilingual models and hence create a filtered version of the Kreol Morisien part\footnote{As mentioned in Section~\ref{sec:prior}, we filter to remove the KreolMorisienMT test set sentences from CreoleM2M's training data.} of CreoleM2M's training data. We use this to train separate tokenizers of 16,000 subwords for Kreol Morisien and English. One tokenizer is with this filtered version alone, and one is with a combination of the filtered version and the training data of KreolMorisienMT.

\begin{table}[t]
\small
\resizebox{\columnwidth}{!}{%
\begin{tabular}{llllll}
\toprule
\textbf{Model} & \textbf{Source} & \textbf{Target} & \textbf{\# Lines} & \textbf{BLEU} & \textbf{chrF} \\ \hline
\multirow{3}{*}{OPUS} & es & ht & 102 & 12.1 & 32.9 \\
 & fr & ht & 1,503 & 11.8 & 33.5 \\
 & en & ht & 1,559 & 14.7 & 35.8 \\ \midrule
\multirow{2}{*}{CreoleM2M} & en & ht & \multirow{2}{*}{1,559} & 22.0 & 43.9 \\
 & ht & en &  & 18.6 & 38.1 \\
\bottomrule
\end{tabular}
}

\caption{Performance of OPUS models (opus-mt-en-ht, opus-mt-es-ht, opus-mt-fr-ht) on our MIT-Haiti Corpus benchmarks, as well as the results of decoding the MIT-Haiti benchmarks using the fine-tuned CreoleM2M Haitian Creole models. }
\label{tab:mit-ayiti}
\end{table}

\begin{table*}[th]
\small
\centering
\begin{tabular}{llrrrr}
\toprule
\multirow{2}{*}{\textbf{Data}} & \multirow{2}{*}{\textbf{Model}} & \multicolumn{2}{c}{\textbf{BLEU}} & \multicolumn{2}{c}{\textbf{chrF}} \\ \cline{3-6}
 &  & \textbf{mfe-eng} & \textbf{eng-mfe} & \textbf{mfe-eng} & \textbf{eng-mfe} \\\hline
\citet{dabre-sukhoo-2022-kreolmorisienmt} & Scratch & 11.1 & 11.5 & - & - \\
\citet{dabre-sukhoo-2022-kreolmorisienmt} & mBART-50-MT-FT & 24.9 & 22.8 & - & - \\\midrule
CreoleM2M & Scratch & 16.1 & 11.5 & 38.0 & 37.1 \\
CreoleM2M+KreolMorisienMT & Scratch & 20.5 & 16.9 & 42.8 & 41.1 \\
CreoleM2M & mBART-50-MT-FT & 22.1 & 18.9 & 44.6 & 44.4 \\
CreoleM2M+KreolMorisienMT & mBART-50-MT-FT & 25.7 & 24.7 & 47.8 & 48.2 \\
\bottomrule
\end{tabular}
\caption{Results on the KreolMorisienMT test sets by using CreoleM2M training data, in addition with the training data in KreolMorisienMT.}
\label{tab:kreolmorisienmtreprod}
\end{table*}

Table~\ref{tab:kreolmorisienmtreprod} contains results for the test set of KreolMorisienMT. We compare our models trained from scratch and fine-tuning against those of \citet{dabre-sukhoo-2022-kreolmorisienmt}. The most important thing to note is that our scratch models are overwhelmingly better than corresponding models by \citet{dabre-sukhoo-2022-kreolmorisienmt}. In fact, we see gains of up to 9.4 BLEU. On the other hand, the filtered CreoleM2M data when used for fine-tuning, despite its size, does not lead to a model that surpasses \citet{dabre-sukhoo-2022-kreolmorisienmt}'s corresponding model that is fine-tuned on a much smaller KreolMorisienMT training dataset. However, by combining both the filtered CreoleM2M and KreolMorisienMT training datasets, we finally surpass \citet{dabre-sukhoo-2022-kreolmorisienmt}'s best results\footnote{\citet{dabre-sukhoo-2022-kreolmorisienmt} do not give chrF scores in their paper and do not release their translations, making it impossible for us to compare chrF scores}.

\paragraph{Other}
We exclude \textbf{PidginUNMT} \cite{Ogueji2019PidginUNMTUN}, as this unlabeled dataset pertains to unsupervised machine translation, and thus cannot be used as gold-standard evaluation data. We also exclude \textbf{WMT11} \cite{callison-burch-etal-2011-findings}, as it was created to help victims of the 2010 earthquake in Haiti, and thus contains sensitive data.  

\section{Discussion and Recommendations}\label{sec:discussion}

\paragraph{Implications for Transfer Learning}
The introduction of \textsc{CreoleVal} marks a significant step forward in bridging the technological divide for Creole languages, in the context of NLP. 
Prior to this work, the scarcity of resources for Creoles made progression of NLP tailored for Creole speakers close to impossible. Now, as shown in Figure~\ref{fig:map}, 28 Creole languages are part of a unified platform, despite previously having limited or no NLP datasets. 
This platform enables researchers and developers to easily include Creoles in pre-existing pipelines, introducing a novel and unique low-resource scenario to NLP.
Given the genealogical ties of many Creoles to (typically) higher-resourced languages\footnote{Some Creoles have strong genealogical ties to lower-resourced languages, such as the Niger-Congo Creoles Lingala, Kikongo-Kituba, Fanakalo, which are related to Bantu languages, and Sango, which is related to Ngbandi.}, we expect this to allow for nuanced experimentation in transfer learning.
In particular, the complex picture of Creoles, including both horizontal and vertical transfer between diverse languages, may offer the key to developing transfer learning techniques which are tuned to encapsulate specific pieces of cross-linguistic knowledge. 
While vocabulary might be transferred from a parent language, syntactic and semantic structures may diverge, challenging conventional transfer learning methods.
Indeed, previous work has shown the difficulties of straight-forward transfer learning techniques from ancestor languages \cite{lent-etal-2022-ancestor}.
We suggest that the success of transfer learning in this new domain relies on in-depth understanding of the structural and contextual intricacies of each individual Creole language, rather than a simplistic reliance on their parent languages.
Moreover, we believe that work to this end has the potential to  improve transfer learning methodology, as it will help researchers gain a broader understanding of the capabilities and limitations of transfer learning.
Finally, beyond strict transfer learning, we also expect cultural adaptation to be a significant challenge for the future, for which \textsc{CreoleVal} provides a benchmark.

\paragraph{Further Resource Development}
While \textsc{CreoleVal} opens for straightforward inclusion of a set of Creole languages in NLP pipelines, we are still limited to textual data.
While this is an important contribution which may lead to a more even playing field in terms of language technologies, it is not enough to focus on this modality.
Considering the fact that many Creoles are exclusively \textit{spoken} languages indicates that a focus on speech resource development is an important next step.

\paragraph{Recommendations}
For future work on Creole languages, be it in the context of experimentation on \textsc{CreoleVal}, or on further resource development, we recommend the following:
\begin{enumerate}
    \item Engage with language communities. 
    When languages are limited in resources, it is critical that any new additional resources are allocated to efforts that will benefit the communities using the language in question \cite{bird-keynote-emnlp-2021}. 
    For Creoles, a concrete starting point is to reach out to experts, as discussed by \citet{lent-etal-2022-creole}.
    \item Keep in mind contextual factors such as domain and culture. 
    Direct translations in narrow domains are likely to introduce cultural biases, which may render language technology less relevant to potential end-users \cite{hershcovich-etal-2022-challenges}. 
    When it is not possible to gather naturally occurring language data, we echo similar recommendations by others for culturally sensitive translations \cite{roemmele2011choice}.
\end{enumerate}

\section{Conclusion}

In this work, we have addressed the absence of Creole languages from contemporary NLP research by introducing benchmarks and baselines for a total of 28 Creole languages.
We argue that this omission in previous work has hindered the progress of NLP technologies tailored to Creole-speaking populations, in addition to preventing research communities from exploring the unique linguistic situations of this diverse group of languages.
With the introduction of \textsc{CreoleVal}, we have made a significant step towards bridging the gap between Creole languages and other low-resource languages in NLP.
We hope that the public release of our datasets and trained models will serve as an invitation to further research in this relatively unexplored domain, and expect that NLP and computational linguistics research stand to gain significantly from embracing the linguistic and cultural diversity embodied in this group of languages.

\section*{Limitations}
Although we are the first to create NLU and NLG benchmarks for up to 28 Creoles, we note the following limitations.

\paragraph{Limited domain diversity} While we were able to collect reasonably large parallel corpora for Creole MT, the data itself belongs to the religious domain and thus might not be extremely useful in a general purpose MT setting. Controversially, the Bible and other religious texts may be considered colonialist by some communities, as these texts may be used to "\textit{provoke a culture change in these communities}" \cite{mager-etal-2023-ethical}. However, works in domain adaptation \cite{chu-etal-2017-empirical, imankulova-etal-2019-exploiting} have shown that even a small amount of in-domain corpus may be sufficient for adaptation to other domains.

\paragraph{Mixture of data quality} 
In this work, we put forth and experiment with a combination of higher and lower quality data, the latter coming from the religious domain.
Works in NLP have long relied on religious texts for truly low-resource languages, which often have no other available data \cite{agic-etal-2015-bit, agic-etal-2016-multilingual}. However, the use of such data comes with concerns over data quality, as such texts are often written by foreign missionaries, they cannot be considered strictly representative of the language as used by native speakers \cite{nida1945linguistics}. 
While the inclusion of religious data is still a common necessity in the realm low-resource NLP, the addition of our higher quality data for Creoles ensures that future works will have a wider variety of resources to evaluate their systems, than previously available. 
Moreover, when sourcing data from domains like Wikipedia, we involve speakers and cross-reference linguistic grammars, leading us to exclude several languages due to quality issues, such as Pitkern.

\paragraph{Lack of reliable monolingual corpora sources} Unlike resource-rich languages like English, French, and Hindi, finding monolingual corpora for Creoles is extremely difficult. One reason for this is the historic lack of interest in research on Creoles in NLP. The lack of monolingual corpora also inhibits the development of LLMs for Creoles, however even a tiny amount may be helpful for expanding existing LLMs, as shown by \citet{yong-etal-2023-bloom}.

\paragraph{Language identification tools} A possible reason for the difficulty in obtaining Creole corpora from the web is that there are extremely limited language identification (LID) \cite{baldwin-lui-2010-language} tools for Creoles, and thus identifying Creole content in CommonCrawl\footnote{\url{https://commoncrawl.org/}} is also very difficult. Developing LID tools for Creoles will be an important future work \cite{kargaran-etal-2023-glotlid}.  

\paragraph{Modality} Many Creoles are spoken and not written, therefore text-based NLP might not be suited for them. This motivates branching out into speech-to-text (automatic speech recognition, speech translation) and speech-to-speech (translation) research.

\section*{Acknowledgments}
HL, YC, MF, EP, HEH, and JB are funded by the Carlsberg Foundation, under the \textit{Semper Ardens: Accelerate} programme (project nr. CF21-0454).
EL is funded by the Google Award for Inclusion Research program (awarded to HL and JB for the ``CREOLE: Creating Resources for Disadvantaged Language Communities'' project).
For KT and MDL, the computational resources and services used were provided by the VSC (Flemish Supercomputer Center), funded by the Research Foundation - Flanders (FWO) and the Flemish Government - department EWI.
MIT-Haiti is, in the main, internally funded by grants from Jameel World Education Lab\footnote{\url{https://www.jwel.mit.edu/}} (for MDG).
Some experiments were enabled by resources provided by the National Academic Infrastructure for Supercomputing in Sweden (NAISS) at Chalmers partially funded by the Swedish Research Council through grant agreement no.\ 2022-06725 (for MB).
The translations of the MCTest dataset were funded by the European Union’s Horizon 2020 research and innovation programme under the Marie Skłodowska-Curie grant agreement No 801199 (for HL) \euflag. 
For the translation of MCTest into Mauritian Creole, we thank Hugues Marianne for his diligent work.
For additional help with the verification of the relation classification datasets, we are deeply grateful to Paweł Kornacki, Krzysztof Kosecki, Gracie Rhule, Tayvia Henry, Dahlia Richards-White and Humroy White, Shanice Carr, Ghawayne Calvin, Xander Gregory, and April Joy A. Molina. Finally, we would like to thank Mike Zhang for his comments on our manuscript, as well as the TACL reviewers and action editor for their indispensable feedback.

\section*{Contributions}
We use CRediT (Contributor Roles Taxonomy \url{https://credit.niso.org}) to note the different roles undertaken by the authors: \\
\textbf{Conceptualization}	AS, JB, HL; \\
\textbf{Data Curation}	HL, RD, YC, RAA, AE, CM, MF, HEH, EL, PB; \\
\textbf{Formal Analysis} HL, KT, RD, YC, MF, EP, LZ, DK, MB, LG; \\
\textbf{Funding Acquisition} JB, HL; \\
\textbf{Investigation}	HL, RD, MDL, DH, MDG, AS, JB; \\
\textbf{Methodology \& Software} HL, KT, RD, YC, MF, EP, LZ, HEH, DK, MB,LG; \\
\textbf{Project Administration} HL; \\ 
\textbf{Resources}	AS, JB, RD, MB, LG, MDG; \\
\textbf{Validation}	HL, KT, RD, YC, LZ, MB; \\
\textbf{Writing} HL, KT, RD, YC, MF, EP, LZ, DK, MB, MDL, DH, MDG, JB. \\

\bibliography{tacl2021}
\bibliographystyle{acl_natbib}

\newpage

\appendix

\section{Relation Classification}\label{appendix:rel_class}
Here, we thoroughly describe our steps to create the relation classification datasets, from data collection, to annotation and verification. This discussion is intended to provide details for exact replication of the work described in the paper, for creating these datasets.
For an overview, our methodology consisted of the following steps:

\begin{enumerate}
  \item Collecting and cleaning data from Wikipedia dumps, and performing automatic entity linking. 
  \item Clustering sentences which belong to the same latent template (i.e., the sentences express the same relation, as evidenced by an exact or near-exact overlap in the text, with the only differences being the entities; more details are provided in Appendix~\ref{appendix:templates}).
  \item Manually verifying and correcting any mistakes from the automatic entity-linking.
  \item Manually annotating the relation expressed in the sets of utterances (as grouped by the latent templates) and its associated Property in Wikidata.
  \item Validating that the annotated triples indeed exist in Wikidata; sentences where the triples did not exist in Wikidata (due to gaps in the knowledge base) were thrown out.
 \item Manually checking the correcting the annotated sentences to ensure that the samples truly reflect real-world usage of the language. 
  \begin{enumerate}
    \item A manual verification of each dataset was performed by a speaker of each Creole. Each sentence was assessed, and speakers made corrections to the grammar or spelling, as they saw fit. Whenever possible, an additional speaker was asked to double-check these changes. 
    \item Complementing the above step, a manual verification of the datasets is conducted using published linguistic grammars for the relevant language, to help identify potential issues in the data.
    \item A final re-verification of the entity tagging and property labels was conducted, to ensure that any corrected sentences were still properly annotated. 
  \end{enumerate}
\end{enumerate}

For steps 1-4, we produced datasets for: Bislama, Chavacano, Haitian Creole, Jamaican Patois, and Pitkern, and Tok Pisin. However at step 5, the triples for Haitian Creole were not validated by the Wikidata and thus this dataset was discarded. Here, simple triples like (apple, is\_a, fruit) were missing from the knowledge graph. Additionally at step 6, the Pitkern samples failed to conform with the description of the language detailed in the grammar, and was also excluded from this work. Ultimately, this resulted in high-quality relation classification evaluation data for 4 of the 9 Creole Wikipedias we started with: Bislama, Chavacano, Jamaican Patois, and Tok Pisin.

\subsection{Data Collection and Annotation}
We first clean the data and perform automatic entity linking and filtering, in order to facilitate the process of manual annotation. First, we preprocess the Wikipedia dumps by removing unnecessary HTML with Beautifulsoup and tokenization with Spacy. We then automatically label entities and link them to Wikidata, a process known as entity linking, first by linking tokens with existing Wikipedia hyperlinks within the text, and then attempt to label any remaining entities without hyperlinks by leveraging OpenTapioca. Before any manual annotation over these examples, we then attempt to automatically group sentences by latent templates, so that sentences can be annotated in groups, allowing us to identify and annotate the correct relationship between the entities, as expressed in the sentences (see “Latent Templates”, below). To this end, we perform automatic clustering over the sentences using first fuzzy string matching with partial token sort ratio, and thereafter affinity propagation, in hopes that utterances sharing templatic spans of text will be clustered together. The result is a large set of clusters, each containing a number of utterances that are at least somewhat similar. In order to refine these clusters further, we first rank the clusters by the longest common string therein, and we then discard clusters below a certain threshold of similarity, as we can assume the sentences do not belong to the same latent template.
Finally, with the highest-scoring clusters of entity-linked sentences, the authors perform a manual annotation of entities and relations.

\subsection{Latent Templates}\label{appendix:templates}

In Section~\ref{sec:rel_class}, we mention the latent templates that the sentences belong to, and how these templates enable more confident manual annotation. To clarify this, we will show some examples of latent templates, and how we map this to Wikidata Properties (i.e. relations) and entities. Note that samples were clustered by latent templates \textit{before} validation and correction by the Creole language speakers, so the provided examples below do not represent the finalized dataset. Consider the following \textcolor{darkblue}{entity-tagged} sentences in Bislama:

\begin{itemize}[noitemsep,topsep=1pt]
    \item \textcolor{darkblue}{Mongolia} i kaontri long \textcolor{darkblue}{Esia}.
    \item \textcolor{darkblue}{Fiji} hem i wan kaontri long \textcolor{darkblue}{Pasifik}.
    \item \textcolor{darkblue}{Jemani} i kaontri long \textcolor{darkblue}{Yurop}.
    \item \textcolor{darkblue}{Bukina Faso} i kaontri long \textcolor{darkblue}{Afrika}.
    \item \textcolor{darkblue}{Kanada} i wan kaontri blong \textcolor{darkblue}{Not Amerika}.
\end{itemize}

When we look at these sentences as a group (i.e. a cluster), we can see there is a latent template of \textcolor{gray}{[[ABC]] (hem) i (wan) kaontri (b)long [[XYZ]]}. All sentences in the cluster belong to this latent template, albeit with some minor variations, which are later inspected and assessed in detail during the validation stage by a speaker of Bislama, and additional with a cross-reference against a linguistics grammar documenting the language.  

Moving on, for the \textbf{entities} themselves, we can identify the Wikidata Qcode in 2 ways:

\begin{enumerate}
    \item The entities (e.g. \textcolor{darkblue}{Mongolia}, \textcolor{darkblue}{Pasifik}) were already hyperlinked in the Wikipedia article, which means we have a URL, from which we can get the gold entity Q-code.
    \item The entities are Named Entities with spelling clearly influenced by English, and we can make an educated guess about the meaning.
\end{enumerate}

Thus from the template and entities, we can now consider the \textbf{relation} between the entities:

(\textcolor{darkblue}{Mongolia} is to \textcolor{darkblue}{Asia}) as (\textcolor{darkblue}{Fiji} is to \textcolor{darkblue}{Pacific}) as (\textcolor{darkblue}{Germany} is to \textcolor{darkblue}{Europe}) as (\textcolor{darkblue}{Canada} is to \textcolor{darkblue}{North America}) and (\textcolor{darkblue}{Burkina Faso} is to \textcolor{darkblue}{Africa}).

For all of these entity pairs, to a human annotator, it is clear that the relationship is \textcolor{gray}{[[COUNTRY]] is in [[CONTINENT]]}. Thus we can annotate the Wikidata Property as P30: "continent of which the subject is a part".

Finally, we can automatically verify our triples (entity1, Property, entity2) against the Wikidata knowledge graph. We remove any sentences where the triple was not in the knowledge graph. This unfortunately removes correct data points, where there is simply a gap in the knowledge graph; for example, the Haitian dataset was removed for this reason, as Wikidata missed simple cases like (apple, is\_a, fruit). But importantly, it also is a sanity measure of our annotation method performed by the authors, which at times required educated guesswork about the meaning of an entity, as non-native speakers, when the entity was not already hyper-linked. Presumably, if we incorrectly annotated an entity, the triple will not exist in the knowledge graph, and thus be removed. Imagine that we had incorrectly annotated [[\textcolor{darkblue}{Kanada}]] (from the sentence [[\textcolor{darkblue}{Kanada}]] i wan kaontri blong [[\textcolor{darkblue}{Not Amerika}]].) to be the language \textit{Kannada} (Q33673)), rather than the country \textit{Canada} (Q16). The triple (Kannada\_language, "continent of which the subject is a part", North America) would certainly not exist in Wikidata, and thus the entire annotated example would be removed. Yet (Canada, "continent of which the subject is a part", North America) is indeed in the knowledge base, so we can be confident in our annotation. Again, having samples listed together in groups by latent templates also makes us more certain of the meaning.

Here are some more examples of latent templates in the data, and the expressed relations:

\textbf{Chavacano}\\
Latent template: \textcolor{gray}{[[PERSON]] is a [[SINGER]]}.\\
Property P106: "occupation of a person" \\ 
Examples: 

\begin{itemize}[noitemsep,topsep=1pt]
    \item \textcolor{darkblue}{Billie Eilish} es un \textcolor{darkblue}{cantante}
    \item \textcolor{darkblue}{Sopho Khalvashi} es un \textcolor{darkblue}{cantante}
    \item \textcolor{darkblue}{Juanes} es un \textcolor{darkblue}{cantante} de Colombia de pop.
    \item \textcolor{darkblue}{Nina Sublatti} (Sulaberidze) es un \textcolor{darkblue}{cantante}
    \item \textcolor{darkblue}{Nini Shermadini} es un \textcolor{darkblue}{cantante}
\end{itemize}

\textbf{Jamaican Patois}\\
Latent template: \textcolor{gray}{[[CITY]] is the capital of [[COUNTRY]]}\\
Property P1376: "capital of" \\
Examples: 
\begin{itemize}[noitemsep,topsep=1pt]
    \item \textcolor{darkblue}{Sofiya} a di kiapital fi \textcolor{darkblue}{Bulgieria}.
    \item \textcolor{darkblue}{Broslz} a di kiapital fi \textcolor{darkblue}{Beljiom}.
    \item \textcolor{darkblue}{Ruom} a di kyapital fi \textcolor{darkblue}{Itali}.
    \item \textcolor{darkblue}{Masko} a di kyapital fi \textcolor{darkblue}{Rosha}.
    \item \textcolor{darkblue}{Atenz} a di kyapital fi \textcolor{darkblue}{Griis}.
\end{itemize}

\subsection{Validation and Corrections}
After the manual validation of the entity tagging and relation labeling, as well as automatic validation of triples in Wikidata, we work with speakers to further validate and correct the sentences as needed, as sentences sourced from Wikipedia can often be of poor quality for lower-resource languages \cite{kreutzer-etal-2022-quality}. In conjunction with the validation performed by speakers, we also check published linguistic grammars for these languages, to ensure that our published datasets constitute the up-most quality. 

\paragraph{Validation and Corrections by Speakers}
For Bislama, Chavacano, Jamaican Patois, and Tok Pisin, we collaborated with at least one speaker of the language to validate and correct the annotated samples. 
Here, our speakers are either semi-native speakers (i.e., they grew up using the language), or professional linguists who live in the pertinent community and speak the language on a daily basis. Indeed, as many Creoles exist as a lingua-franca in multilingual communities, there are not always ``native speakers'', in the sense that the Creole will be their mother tongue \cite{lent-etal-2022-creole}. We provide details and discussion on the validation and corrections made for each language below: 

\begin{itemize}[noitemsep,topsep=1pt]
    \item \textbf{Bislama}: the samples were corrected by one speaker. Overall, the speaker found that some sentences were completely correct, fluent Bislama, with minor spelling errors. Almost all sentences were understandable, but many contained specific grammatical errors or contained many spelling errors. Only a few sentences were completely wrong, and corrected accordingly, to capture the meaning of the annotated triple. The major grammatical errors involved missing prepositions, incorrect usage of articles, or incorrect verb tense. 
    \item \textbf{Chavacano}: the samples were corrected by one speaker, and further validated by a second. Here, the sentences in Wikipedia were determined not to be Chavacano, but rather an approximation of Spanish. As the intended meaning of the utterances was still clear, the speaker produced new utterances in Chavacano, to correctly capture the intended meaning with the tagged entities and labeled relation. 
    \item \textbf{Jamaican}: the samples were corrected by one speaker, and further validated by six others. The spelling and grammar of the Wikipedia sentences was found to be greatly divergent from real-world Jamaican, and thus not representative of the language. Specifically the orthography did not match what is used by Jamaican speakers, and there were a number of grammatical constructions that would not be used by native speakers. To remedy this, the speaker produced new utterances in Jamaican, to correctly capture the intended meaning with the tagged entities and labeled relation. 
    \item \textbf{Tok Pisin}: the samples were validated by two speakers, who noted that while the data is correct, it is distinctly representative of the urban variety of the language (\textit{Tok Pisin bilong taun}), which can vary greatly from the rural variety (\textit{Tok Pisin bilong ples}). Thus for future work, collecting and annotating samples that capture a wider spectrum of Tok Pisin will be key for expanding language technology to this language. 
\end{itemize}

After all manual corrections were made, we conduct an additional round of manual validation, to ensure that the entity tagging and relation labels were still correct. 

One common thread across all languages involved spelling, as many Creoles do not have strictly observed orthography. For example, for lesser-known named entities, there is likely to be great variation across speakers, in whether they default to English spelling, or rather attempt to represent the word according to their pronunciation. This issue highlights an area of future work, for extending Creole language datasets to capture a wider variety of voices and approaches to spelling. To this point, some speakers chose to add limitation variation across their corrections of the data. For example, in the Bislama dataset, there can be found variation in constructions combing the 3person-singular pronoun and the predicate marker \textit{i}. 

Finally, while we did not have funds to pay the speakers for their assistance in this work, the speakers were invited to join the project as co-authors of this work, or otherwise be thanked by name in the Acknowledgements, per their preference. We believe no speakers were harmed in this process, and we are deeply grateful for their collaboration in this work. 

\paragraph{Validation through Linguistic Grammars}
Full documentation of our grammar check has been submitted as supplementary material alongside this manuscript, for inspection by the reviewers. As we cite directly from published books, copyright prevents us from making our grammar check public. For Bislama we referred to \citet{crowley2004bislama},
for Chavacano we referred to \citet{lipski2007zamboangueno}, and 
for Jamaican Patois we primarily referred to \citet{patrick2014jamaican}, but also referenced others \cite{durrleman2008syntax, patrick2004jamaican, bailey1966jamaican}.
For Pitkern we referred to \citet{muhlhausler2020pitkern},
and finally for Tok Pisin we referred to \citet{eberl2019innovation}. 
Amongst all of these languages, Pitkern was the only case where the Wikipedia data failed to meet the description of language, and was thus removed.

\section{Machine Translation: Creole M2M}\label{appendix:mt}
\subsection{Dataset Statistics}

Table \ref{tab:statsm2m} shows the statistics of the training set of the CreoleM2M dataset, spanning 26 Creoles originating from one or more of 8 parent (ancestor) languages. 
We give the number of lines, and number of words on the source (Creole) and target (English) sides.

\begin{table*}[!ht]
    \centering
        \resizebox{ \textwidth}{!}{
    \begin{tabular}{llllll}
    \hline
        \textbf{Pair} & \textbf{Creole} & \textbf{Ancestor(s)} & \textbf{\#Lines} & \textbf{\#Words-Source} & \textbf{\#Words-Target} \\ \hline
        \textbf{hwc-eng} & Hawaiian Pidgin & English & 4,366 & 144,281 & 102794 \\ 
        \textbf{acf-eng} & Saint Lucian Creole & French & 4,889 & 135,006 & 115176 \\ 
        \textbf{gul-eng} & Gullah & English & 4,889 & 153,823 & 115,176 \\ 
        \textbf{icr-eng} & San Andrés–Providencia Creole & English & 4,889 & 151,372 & 115,176 \\
        \textbf{mbf-eng} & Malay Baba & Malay & 4,889 & 107,234 & 115,176 \\ 
        \textbf{ktu-eng} & Kituba & Kikongo & 4,889 & 103,577 & 115,176 \\ 
        \textbf{jam-eng} & Jamaican Creole & English & 5,012 & 206,692 & 168,134 \\ 
        \textbf{tcs-eng} & Torres Strait Creole & English & 6,350 & 198,593 & 152,642 \\ 
        \textbf{mkn-eng} & Kupang & Malay & 6,422 & 214,390 & 153,596 \\ 
        \textbf{cbk-eng} & Chavacano Creole & Spanish & 7,071 & 182,859 & 127,090 \\ 
        \textbf{bzj-eng} & Belizean & English & 12,085 & 262,496 & 218,526 \\ 
        \textbf{rop-eng} & Australian Kriol & English & 27,617 & 832,308 & 703,888 \\ 
        \textbf{pcm-eng} & Nigerian Pidgin & English & 28,267 & 523,916 & 459,266 \\ 
        \textbf{srm-eng} & Saramaccan Language & English, Portuguese & 39,640 & 973,176 & 627,273 \\ 
       \textbf{kri-eng} & Sierra Leonean Creole & English & 47,673 & 1,039,743 & 760,699 \\
        \textbf{djk-eng} & Aukan & English & 58,108 & 1,487,156 & 1,015,311 \\ 
        \textbf{tdt-eng} & Tetun Dili & Portuguese & 118,461 & 2,209,118 & 1,923,333 \\ 
        \textbf{mfe-eng} & Mauritian Creole & French & 189,877 & 3,549,493 & 3,014,530 \\ 
        \textbf{hat-eng}& Haitian Creole & French & 208,772 & 4,132,691 & 3,322,288 \\ 
        \textbf{crs-eng} & Seychellois Creole & French & 220,861 & 3,984,410 & 3,750,620 \\ 
        \textbf{sag-eng} & Sango & Ngabandi, French & 260,853 & 6,089,066 & 4,246,373 \\ 
        \textbf{pis-eng} & Pijin & English & 277,378 & 4,783,222 & 4,458,132 \\ 
        \textbf{pap-eng} & Papiamento & Spanish & 396,092 & 7,297,575 & 6,384,282 \\ 
        \textbf{tpi-eng} & Tok Pisin & English & 399,486 & 8,365,958 & 6,334,237 \\ 
        \textbf{bis-eng} & Bislama & English & 488,393 & 10,751,097 & 7,903,431 \\ 
        \textbf{srn-eng} & Sranan Tongo & English & 583,746 & 13,450,377 & 9,911,997 \\\hline
        \textbf{Total} & - & - & 3,410,975 & 71,329,629 & 56,314,322 \\ \hline
    \end{tabular}}
    \caption{Statistics of the training set of the CreoleM2M dataset.}
\label{tab:statsm2m}

\end{table*}

\section{Overview}\label{appendix:overview}
\begin{table*}[t]
\centering
\resizebox{\textwidth}{!}{%
\begin{tabular}{llp{2.5cm}lp{2cm}lrr}
\toprule
\textbf{Task}                & \textbf{Dataset}       & \textbf{Language (ISO-638-3)}                                                                       & \textbf{Metric}          & \textbf{License}  & \textbf{Domain} & \textbf{Total Sent.} & \textbf{Total words} \\ \midrule
\multirow{2}{*}{MC}                           & \multirow{2}{*}{CreoleVal MC}           & hat-dir, hat-loc, mfe                                                                               & Acc                      & Microsoft License & Education   & 3894                 & 32068                \\ 
\multirow{2}{*}{RC}                           & \multirow{2}{*}{CreoleVal RC}           & bis, cbk, jam, tpi                                                                                     & F1                       & CC0               & WikiDump        & 785                  & 4106                 \\ 
\multirow{8}{*}{MT}                           & \multirow{8}{*}{CreoleVal Religious MT} & bzj, bis, cbk, gul,
hat, hwc, jam, ktu,
kri, mkn, mbf,
mfe, djk, pcm,
pap, pis, acf, icr,
sag, srm, crs, srn,
tdt, tpi, tcs & \multirow{8}{*}{Bleu, chrF}               & \multirow{8}{*}{Copyrighted}       & \multirow{8}{*}{Religion}        & \multirow{8}{*}{64394}                & \multirow{8}{*}{811741}               \\ 
MT                           & CreoleVal MIT-Haiti    & hat                                                                                                 & Bleu, chrF               & CC 4.0            & Education   & 3164                 & 36281                \\ 
Pretraining data             & CreoleVal MIT-Haiti    & hat                                                                                                 & N/A                      & CC 4.0            & Education   & 8281                 & 116444               \\ 
\midrule
\midrule
\multirow{2}{*}{UDPoS}       & Singlish Treebank$\diamond$ \cite{wang-etal-2017-universal}         & singlish                                                                                            & Acc     & MIT               & Web Scrape            & 1200                 & 10989                \\
                             & UD\_Naija-NSC$\diamond$ \cite{caron-etal-2019-surface}           & pcm                                                                                                 &         Acc                 & CC 4.0            & Dialog          & 9621                 & 150000               \\ 
\multirow{2}{*}{NER} & MasakhaNER$\diamond$ \cite{adelani-etal-2021-masakhaner}             & pcm                                                                                                 & Span-F1 & Apache 2.0        & BBC News        & 3000                 & 76063                \\
                             & WikiAnn$\star$ \cite{pan-etal-2017-cross}                & bis cbk hat, pih, sgg, tpi, pap                                                                          &    Span-F1                      & Unspecified           & WikiDump        & 5877                 & 74867                \\ 
\multirow{2}{*}{SA}          & AfriSenti$\diamond$ \cite{muhammadSemEval2023}              & pcm                                                                                & Acc     & CC BY 4.0         & Twitter         & 10559                & 235679               \\
                             & Naija VADER$\star$ \cite{Oyewusi2020SemanticEO}            &   pcm                                                                                                  &  Acc                        & Unspecified           & Twitter         & 9576                 & 101057               \\ 
NLI               & JamPatoisNLI$\diamond$ \cite{armstrong-etal-2022-jampatoisnli}           & jam                                                                                                 & Acc                      & Unspecified               & Twitter, web    & 650                  & 2612                 \\ 
\multirow{3}{*}{SM}                        & \multirow{3}{*}{Tatoeba$\star$ \cite{artetxe-schwenk-2019-massively}}               & cbk, gcf, hat, jam, pap, sag, tpi                                                                         & \multirow{3}{*}{Acc}                      & \multirow{3}{*}{CC-BY 2.0}               & \multirow{3}{*}{General web}     & \multirow{3}{*}{49192}                & \multirow{3}{*}{319719}               \\ 
MT                           & KreolMorisienMT$\diamond$ \cite{dabre-sukhoo-2022-kreolmorisienmt}       & mfe                                                                                                 & Bleu, chrF               & MIT License       & Varied          & 6628                 & 23554                \\ \midrule
                             &                        &                                                                                                     &                          &                   & New:          & 80518               & 1000640 \\
                             &                        &                                                                                                     &                          &                   & Total:          & 176821               & 1995180   \\       \bottomrule   

\end{tabular}%
}
\caption{Overview of the datasets included in \textsc{CreoleVal}. Newly introduced datasets are prefixed with ``CreoleVal''; $\star$ indicates modified and further denoised datasets based on previous works; $\diamond$ indicates inclusion within our benchmark where we provide download and experiment scripts as part of our Github repository, but do not re-package the data itself. Task abbreviations: MC (machine reading comprehension); RC (relation classification); MT (machine translation); UDPoS (universal dependencies part-of-speech tagging); NER (named entity recognition); SA (sentiment analysis); NLI (natural language inference); SM (sentence matching). Note that for SM task, the language format is XXX-eng. For WikiAnn, NaijaVADER and JamPatoisNLI datasets, the licenses were not explicitly stated in corresponding repositories.}
\label{tab:datasetStats}
\end{table*}

\end{document}